\pdfoutput=1

\documentclass[11pt]{article}

\usepackage[final]{coling}

\usepackage{times}
\usepackage{latexsym}

\usepackage[T1]{fontenc}

\usepackage[utf8]{inputenc}

\usepackage{microtype}

\usepackage{inconsolata}

\usepackage{graphicx}

\usepackage{pgfplots}
\pgfplotsset{compat=1.18} 
\usepackage{pgfplotstable} 

\usepackage{amsmath} 
\title{AUEB-Archimedes at RIRAG-2025:\\ Is obligation concatenation really all you need?}

\author{ Ioannis Chasandras $^{1}$, \hspace{0.5mm} Odysseas S. Chlapanis$^{1, 2}$ \and  Ion Androutsopoulos$^{1, 2}$ \\
$^{1}$Department of Informatics, Athens University of Economics and Business, Greece\\
$^{2}$Archimedes/Athena RC, Greece
}

\begin{document}
\maketitle
\begin{abstract}
This paper presents the systems we developed for RIRAG-2025, a  shared task that requires answering regulatory questions by retrieving relevant passages. The generated answers are evaluated using RePASs, a reference-free and model-based metric. Our systems use a combination of three retrieval models and a reranker. We show that by exploiting a neural component of RePASs that extracts important sentences (`obligations') from the retrieved passages, we achieve a dubiously high score (0.947), even though the answers are directly extracted from the retrieved passages and are not actually generated answers. We then show that by selecting the answer with the best RePASs among a few generated alternatives and then iteratively refining this answer by reducing contradictions and covering more obligations, we can generate readable, coherent answers that achieve a more plausible and relatively high score (0.639).
\end{abstract}

\section{Introduction}
The Regulatory Information Retrieval and Answer Generation (RIRAG)\footnote{\url{https://regnlp.github.io/}} shared task focuses on the development of systems that can effectively retrieve relevant information from regulatory texts to generate accurate answers for obligation-related queries. It is divided into two subtasks: \emph{passage retrieval}, where systems identify the ten most relevant passages from regulatory documents, and \emph{answer generation}, which requires synthesizing comprehensive answers from the retrieved passages. 

We participated with three systems and released our code publicly.\footnote{\url{https://github.com/nlpaueb/verify-refine-repass}} Each one of them uses a Rank Fusion \citep{10.1145/3471158.3472233} combination of three retrieval models: BM25 \citep{Robertson1994Okapi}, and two neural domain-specific retrievers, based on a law- and a finance-specific embedding model, respectively. We also apply a neural reranker to the top-N retrieved passages. 

For answer generation, our first system adversarially exploits the evaluation metric of the task, called RePASs, by using one of its neural components. Specifically, we extract important sentences (`obligations') from the retrieved passages 
and then concatenate these sentences to get an `answer'. Even though the produced answers may be incoherent and may not answer the question directly, this system achieves a perfect score, much higher than the score of human experts. The second system extends this approach with an LLM that generates an answer (for each question) by iteratively reformulating (as parts of an answer) the extracted obligations of the previous system.
This results in more readable answers, but performance deteriorates to RePASs scores below those of the challenge's baseline \citep{gokhan2024regnlpactionfacilitatingcompliance}. 

Our third system works by 
a) generating multiple candidate answers and using RePASs to select the best answer, and b) iteratively refining the selected answer by removing contradictions and adding `obligation' sentences that increase RePASs. This  system performs worse than the adversarial (first) system, but much better than the baseline, and the answers are coherent and readable.

\section{Task setup}
\label{sec:task}
\textbf{Dataset:} The dataset of the task consists of train, development, and test sets (22k, 2.8k, 2.7k questions respectively).
Passages are retrieved from a corpus of 40 regulatory documents from the Abu Dhabi Global Markets (ADGM) collection. The task organizers used a separate hidden test set, with 446 questions, to evaluate the participants.

\noindent
\textbf{Evaluation:}\label{sec:repass} Passage retrieval is evaluated using recall@10 and MAP@10. Answer generation is evaluated using 
RePASs, a reference-free metric \citep{gokhan2024regnlpactionfacilitatingcompliance}. To calculate RePASs, \emph{entailment} and \emph{contradiction} scores are obtained by comparing each sentence of the retrieved passages (used as premises) with each sentence of the generated answer (hypothesis) using an NLI model. For each generated sentence (of the answer), the highest probabilities for entailment and contradiction (comparing to retrieved sentences) are selected, 
and the scores are averaged over all the sentences of the answer.
Additionally, \textit{obligation}-sentences are extracted from the retrieved passages using a LegalBERT model \citep{chalkidis-etal-2020-legal} fine-tuned on a synthetic dataset \citep{gokhan2024regnlpactionfacilitatingcompliance}. For an obligation to be considered \textit{covered} by the generated answer, a sentence of the answer must entail the obligation-sentence with a confidence above a certain threshold, according to another NLI model. 

\section{Passage retrieval} \label{sec:retrieval}

All three of our systems use the same passage retrieval, which improves upon the baseline retrieval system of the shared task \citep{gokhan2024regnlpactionfacilitatingcompliance} in three ways: a) we use domain-specific neural retrieval models,
b) we extend the Rank Fusion approach \citep{10.1145/3471158.3472233}
to include three models instead of two, and c) we use a reranker.

\subsection{Retrieval models}
We experiment with BM25 \citep{Robertson1994Okapi} and three of the best\footnote{MTEB-law: \url{https://huggingface.co/spaces/mteb/leaderboard?task=retrieval&language=law}}
text embedding models: \texttt{text-embedding-3-large} (OL3) from \texttt{OpenAI} \citep{neelakantan2022textcodeembeddingscontrastive}, \texttt{voyage-law-2} (VL2), and \texttt{voyage-finance-2} (VF2) from \texttt{Voyage}.\footnote{\url{https://docs.voyageai.com/docs/embeddings}} The OL3 embedding model is only used for comparison; it is not included in our final systems, because domain-specific embedding models worked better.
 We also use the \texttt{voyage-rerank-2} reranker.

\subsection{Rank Fusion}

The task combines the financial and legal domains, which motivates using two domain-specific neural retrievers.
Also, according to \citet{10.1145/3471158.3472233}, BM25 should be fused with neural retrievers, because it captures exact term matching better.
Hence, we expand Rank Fusion
to handle three retrievers instead of two, as follows.
\begin{equation}
f(p) = 
a \hat{s}_x(p) + 
b \hat{s}_y(p) + 
(1 - (a + b)) \hat{s}_z(p)
\end{equation}
\noindent
Here $p$ is a retrieved passage, $a$ and $b$ are fusion weights, and $\hat{s}_x(p), \hat{s}_y(p), \hat{s}_z(p)$ are the normalized relevance scores of  the three fused retrievers.

\subsection{Experimental results for retrieval}
We conduct three experiments on the public test set. In Table~\ref{tab:retrieval-main}, we compare the scores of the four single retrieval models. We see that the domain-specific voyage-law-2 (VL2) and voyage-finance-2 (VF2) perform better than BM25 and the generic OL3. 

\begin{table}[h!]
\centering
\begin{tabular}{|c|c|c|}
\hline
\textbf{Model}               & \textbf{Recall@10} & \textbf{MAP@10} \\ \hline
BM25 & 0.6994           & 0.5584          \\ \hline
OL3& 0.7385            & 0.5736          \\ \hline
VL2& 0.7705            & 0.6275          \\ \hline
VF2& \textbf{0.7895}            &\textbf{ 0.6559 }         \\ \hline
\end{tabular}
\caption{Comparison of single retrieval models.}
\label{tab:retrieval-main}
\end{table}

In the second experiment (Table~\ref{tab:rank_fusion_models}), we compare Rank Fusion configurations, again on the public test set. The newly introduced triple Rank Fusion, with BM25, VL2 and VF2, is the best. The values of $a,b$ were selected by trying a few combinations.

\begin{table}[h!]
\centering
\setlength{\tabcolsep}{4pt} 
\begin{tabular}{|l|c|c|c|c|c|}
\hline
\textbf{Rank Fusion} & \textbf{a} & \textbf{b}& \textbf{R@10} & \textbf{M@10} \\ \hline
BM25, OL3& 0.30   & -             & 78.9 & 65.0 \\ \hline
VL2, VF2 &0.40         & -   & 79.4 & 66.0  \\ \hline
BM25, VL2 & 0.25                & -       & 79.9 & 66.5  \\ \hline
BM25, VF2   &0.30              & -      & 80.4 & 67.6  \\ \hline
BM25, VL2, VF2 & 0.25&  0.2 & \textbf{81.1 }& \textbf{69.0} \\ \hline
\end{tabular}
\caption{Comparison of Rank Fusion configurations.}
\label{tab:rank_fusion_models}
\end{table}

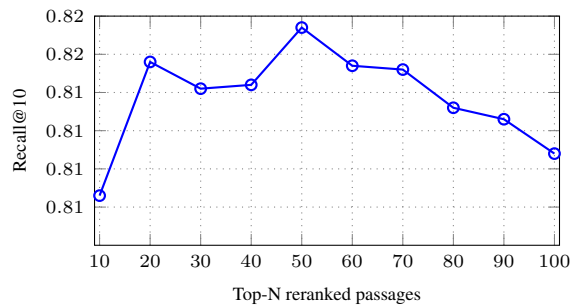
\begin{figure}[h!]
\centering
\begin{tikzpicture}
\begin{axis}[
    width=\columnwidth, 
    height=0.6\columnwidth, 
    xlabel={Top-N reranked passages},
    ylabel={Recall@10},
    xmin=9, xmax=101,
    ymin=0.810, ymax=0.816,
    xtick={0, 10, 20, 30, 40, 50, 60, 70, 80, 90, 100},
    ytick={0.811, 0.812, 0.813, 0.814, 0.815, 0.816},
    grid=both,
    grid style={dotted, gray},
    legend pos=south east,
    legend style={font=\tiny}, 
    ticklabel style={font=\scriptsize},
    label style={font=\scriptsize},
    title style={font=\footnotesize}
]

\addplot[
    color=blue,
    mark=o,
    thick,
    nodes near coords,
    point meta=explicit symbolic 
] coordinates {
    (10, 0.8113)
    (20, 0.8148)
    (30, 0.8141)
    (40, 0.8142)
    (50, 0.8157)
    (60, 0.8147)
    (70, 0.8146)
    (80, 0.8136)
    (90, 0.8133)
    (100, 0.8124)
};

\end{axis}
\end{tikzpicture}
\vspace*{-2mm}
\caption{Recall@10 scores of our best retriever (Rank Fusion of BM25, VL2, VF2) when  reranking the top-$N$ retrieved passages, for different $N$ values.}
\label{fig:rerank}
\end{figure}

In the third experiment (Fig.~\ref{fig:rerank}), we investigate the effect of reranking the top-$N$ retrieved passages, for different $N$ values, by computing Recall@10 on the public test set. 
The best value is $N=50$.

Our final retrieval model is a triple Rank Fusion model (BM25, VL2, VF2) with reranking (\texttt{voyage-rerank-2}, $N=50$), which  
ranked 4th in the retrieval subtask, achieving 69.4  Recall@10, and 59.4 MAP@10 on the hidden test set.


\section{Answer generation}

The answer generators of this section use our best retriever (Section~\ref{sec:retrieval}, BM25, VL2, VF2, reranker).

\subsection{Preprocessing} \label{sec:preprocessing}
\textbf{Filtering:} We follow  \citet{gokhan2024regnlpactionfacilitatingcompliance}, i.e., we rank the retrieved passages by decreasing relevance scores; we then keep only 
passages that satisfy two conditions: (i) their score must be above a certain \textit{threshold}, and (ii) their score must not fall below the previous passage's score more than \textit{max drop}.
\\
\noindent
\textbf{Extracting obligations:} To obtain obligations from the retrieved passages, 
we use the same fine-tuned LegalBERT model used in RePASs (Section~\ref{sec:task}) for obligation extraction. If a passage does not contain any obligations, we use it as is. 

\subsection{Experimental results for preprocessing}

To select the values of the filtering \textit{threshold} and \textit{max drop} (Section~\ref{sec:preprocessing}), 
we conducted two experiments 
using \texttt{GPT-4o-mini}\footnote{\url{https://openai.com/index/gpt-4o-mini-advancing-cost-efficient-intelligence/}} for answer generation.  
The first experiment shows that the recommended values of $0.70, 0.20$ of \citet{gokhan2024regnlpactionfacilitatingcompliance} are outperformed by $0.90, 0.10$, 
respectively (Table~\ref{tab:filtering}).

\begin{table}[h]
\centering
\begin{tabular}{|c|c|c|}
\hline
\textbf{Threshold} & \textbf{Max Drop}& \textbf{RePASs} \\ \hline
0.70           & 0.20             & 0.4708          \\ \hline
0.75               & 0.05              & 0.5006          \\ \hline
0.80               & 0.05              & 0.5050          \\ \hline
0.85               & 0.15               & 0.5001          \\ \hline
\textbf{0.90 }              & \textbf{0.10}              & \textbf{0.5117}          \\ \hline
\end{tabular}
\caption{Performance of the baseline answer generator for different values of \textit{threshold} and \textit{max drop}, using our best retriever (BM25, VL2, VF2, reranker).}
\label{tab:filtering}
\end{table}
\noindent
The second experiment compared the performance of the task's baseline when (a) the entire retrieved passages were given to the LLM, or (b) only the obligations were given, or (c) only the obligations were given, but with a tailored prompt. No significant difference was noticed between (a) and (b), but (c) was significantly better in RePASs (Table~\ref{tab:preprocess_obligations}), due to the increase in \emph{obligation coverage} and \emph{entailment}, even though \emph{contradiction} was worse. 
All prompts can be found in Appendix~\ref{app:baseline}.

\begin{table}[h]
\centering
\setlength{\tabcolsep}{4pt}
\begin{tabular}{|c|c|c|c|c|}
\hline
\textbf{Context}   & \textbf{RePASs}& \textbf{Obl.} & \textbf{Ent.} & \textbf{Con.}  \\ \hline
Passages           & 0.411             & 0.147 & 0.177              &\textbf{0.090}       \\ \hline
Obligations         & 0.413          & 0.156                       & 0.172              & \textbf{0.090}               \\ \hline
 + prompt        & \textbf{0.512 }   & \textbf{0.278}       & \textbf{0.366}              & 0.109                                        \\ \hline
\end{tabular}
\caption{Performance of the baseline system for different kinds of inputs (entire retrieved passages, obligations only, obligations with tailored prompt).}
\label{tab:preprocess_obligations}
\end{table}

\begin{table*}[t!]
\centering
\begin{tabular}{|c|c|c|c|c|}
\hline
\textbf{System / Group Name} & \textbf{RePASs} & \textbf{Obligation} & \textbf{Entailment} & \textbf{Contradiction} \\ \hline
\texttt{GPT-4o} baseline*    & 0.583           & 0.220               & 0.769               & 0.238                  \\ \hline
Human experts*               & 0.859           & \textbf{1.000}      & 0.837               & 0.260                  \\ \hline
Indic aiDias &\textbf{0.973}           & 0.993               &\textbf{0.987      }         & 0.062                  \\ \hline
Ocean's Eleven & 0.971          & 0.991               & 0.986               & 0.065                  \\ \hline
AUEB NLP Group - NOC                          & 0.947 (0.951)   & 0.951 (0.963)      & 0.986 (0.986)       & 0.096 (0.096)          \\ \hline
AUEB NLP Group - VRR                          & 0.639 (0.646)   & 0.502 (0.524)      & 0.446 (0.446)       & \textbf{0.031} (0.031) \\ \hline
AICOE                        & 0.601           & 0.230               & 0.827               & 0.254                  \\ \hline
AUEB NLP Group - LOC                          & 0.562 (0.568)   & 0.423 (0.439)       & 0.375 (0.375)       & 0.110 (0.110)          \\ \hline
\end{tabular}
\caption{Leaderboard results for Subtask 2. Results computed by ourselves for our systems are shown in brackets. Differences are attributed to using different GPUs. *Scores taken from \citet{gokhan2024regnlpactionfacilitatingcompliance}.}
\label{tab:main-table}
\end{table*}

\subsection{Naive Obligation Concatenation (NOC)} \label{sec:noc}

Our first answer generator (NOC) adversarially exploits the extracted obligations (Section~\ref{sec:preprocessing}). It simply concatenates and outputs them as the `answer'. From the definition of RePASs (Section~\ref{sec:task}), this answer should get an almost perfect obligation score. Additionally, we expect a low contradiction score, as obligations should not conflict.

\subsection{LLM Obligation Concatenation (LOC)}


The answers of NOC (Section~\ref{sec:noc}) do not answer the question directly; they are just excerpts from retrieved passages. To alleviate this, we create a variation of NOC, called LOC: for each extracted obligation, we prompt an LLM (\texttt{GPT-4o-mini}) to answer the given question using this obligation. If the generated answer does not \emph{cover} (Section~\ref{sec:repass}) the original obligation, then the LLM is prompted again, until a certain number of tries $K$ has been reached (we use $ K=3 $). Finally, the per-obligation answers are concatenated to form a complete answer.


\subsection{Verify and Refine with RePASs (VRR)}

Our third answer generator (VRR) first `verifies' the correctness of the answers, then iteratively `refines' them. The first stage (verification) is loosely inspired by self-consistency \citep{wang2023selfconsistency}; it involves the generation of many alternative answers by the LLM and the selection of the one with the highest RePASs score. The selected answer is then iteratively refined by reducing \textit{contradictions} and increasing \textit{obligations}, as explained below.

\subsubsection{Verification step}
In the verification step, we obtain $N$ alternative answers from the LLM (using all the extracted obligations and the question as input) and evaluate them using RePASs. We choose the alternative answer with the best RePASs score. 

\subsubsection{Refinement step} 
\textbf{Contradiction removal:} To remove contradictions: a) we compute the average contradiction score over all the answers (over all the best alternative answers for all questions) across the dataset using the same NLI model as in RePASs, and b) we remove the sentences of the answer that get a contradiction score higher than the average.


\noindent
\textbf{Obligation insertion:} To locate missing obligations, we 
extract obligations from the retrieved passages and the current answer. Obligations from the retrieved passages that are not \textit{covered} (Section~\ref{sec:task}) by the current answer are \textit{missing} obligations. We prompt \texttt{GPT-4o} to insert the missing obligations by correcting a sentence or adding a new one to the current answer (complete prompt in Appendix~\ref{app:obligations_insertion_prompt}).

\subsection{Experimental results for generation}
In the following experiments we use the hidden test set, 
\texttt{GPT-4o-mini} as the generator for LOC, and \texttt{GPT-4o}\footnote{\url{https://openai.com/index/hello-gpt-4o/}} as the generator for VRR.

Table~\ref{tab:main-table} compares the task's baseline and human expert performance, as reported by \citet{gokhan2024regnlpactionfacilitatingcompliance}, to our three submissions (NOC, VRR, LOC) and to the best submissions of the top three competitors. NOC achieves an almost perfect RePASs score ($0.947$), surpassing human experts ($+0.088$). As expected, \emph{obligation} and \emph{contradiction} scores are excellent for the adversarial NOC, but surprisingly \emph{entailment} scores are even better without directly optimizing towards them. Similar results are observed for the methods of the top scoring competitors. However, as already mentioned, NOC's answers are just verbatim sentences from the retrieved passages, which proves that RePASs can easily be deceived. LOC on the other hand, which rewrites the `obligations' using \texttt{GPT-4o-mini}, performs even worse than the baseline model, which shows that RePASs is also very sensitive to the style of the answer.
VRR, which actually generates answers from the retrieved passages, improves upon the task's baseline substantially (+0.056) and ranks first among systems that do not exceed human performance; we suspect that systems with super-human performance may trick the RePASs measure, like our NOC system.


The next experiment (Table~\ref{tab:VRR_ablation}) measures the contribution of the verification and refinement processes of VRR. Both processes are beneficial, but verification's improvement is more important. 

\begin{table}[t]
\centering
\begin{tabular}{|c|c|c|c|c|}
\hline
\textbf{VRR}           & \textbf{RePASs} & \textbf{Improvement} \\ \hline
Baseline (Ours)        & 0.506       & -   \\ \hline
+ Verification       & 0.611       & \textbf{+ 0.105}  \\ \hline
+ Refinement  & \textbf{0.646 }      & + 0.025   \\ \hline
\end{tabular}
\caption{Contribution of VRR stages, using GPT-4o.}
\label{tab:VRR_ablation}
\end{table}


\section{Conclusion}
We introduced three systems for the RIRAG shared task. The retrieval backbone of all systems combined BM25 with two domain specific neural retrievers and a reranker. We achieved a near-perfect score with an adversarial system that exploits the neural model for \textit{obligation} extraction of RePASs, highlighting the difficulty of developing a robust reference-free metric for RAG evaluation. 
Our best non-adversarial system (VRR) first generates multiple alternative answers from the retrieved obligations, selects the alternative answer that maximizes RePASs, then iteratively improves it by maximizing obligation coverage and minimizing contradictions. This system produces coherent answers, and obtains the highest RePASs score among competitors that do not exceed human performance (which may be a sign of gaming RePASs). 

\section*{Limitations}
We demonstrated that reference-free model-based metrics, such as RePASs, used for evaluating Retrieval-Augmented Generation (RAG) systems, can be susceptible to adversarial attacks. Specifically, we showed that it is possible to provide answers that receive a high score from the metric, but may not be useful to non-experts. The attack was tailored to RePASs and a specific domain, and it may not  apply to other domains or metrics.

VRR requires an accurate verifier, such as RePASs, which is not always available. The \textit{obligation extraction} component in RePASs is fine-tuned using a synthetic dataset \citep{gokhan2024regnlpactionfacilitatingcompliance}, which in turn requires a powerful LLM teacher to solve the task with few-shot prompting alone. This is quite rare for hard domain-specific problems.

\section*{Acknowledgments}
This work has been partially supported by project MIS 5154714 of the National Recovery and Resilience Plan Greece 2.0 funded by the European Union under the NextGenerationEU Program. All experiments were done using AWS resources which were provided by the National Infrastructures for Research and Technology GRNET and funded by the EU Recovery and Resiliency Facility.

\bibliography{anthology,custom}

\appendix
\section{Related work}
\textbf{RAG:}
Retrieval-Augmented Generation (RAG) \citep{10.5555/3495724.3496517} systems can help tackle domain-specific problems that RegNLP \citep{goanta-etal-2023-regulation} presents, by incorporating information from large regulatory document collections.

\noindent
\textbf{Verify and Refine:} VRR is loosely inspired by LLM methods that select the best answer from multiple candidates and iteratively refine these answers \citep{wang-etal-2024-self-consistency, 10.5555/3666122.3668141, 10.5555/3666122.3666639, quan-etal-2024-verification}, frameworks like Explanation-Refiner \citep{quan-etal-2024-verification} that use theorem proving to validate and refine explanations, and WizardLM \citep{xu2024wizardlm} that evolves
instruction data to enhance model performance.


\noindent
\textbf{Adversarial attacks:}
Many works implement adversarial attacks that are similar to our NOC system. BERT-ATTACK \citep{li-etal-2020-bert-attack} leverages a pretrained BERT model to deceive other models. \citet{huang-baldwin-2023-robustness} show that popular model-based evaluation metrics for machine-translation are susceptible to inconsistencies when given  adversarially-degraded translations.

\section{Prompts}
For all our prompts we have used \texttt{GPT-4o} to improve them, and then kept those that performed the task better (according to our opinion) in a few (2-3) sample questions. 

\subsection*{Baseline prompt \citep{gokhan2024regnlpactionfacilitatingcompliance}}
\label{app:baseline}

\textit{You are a regulatory compliance assistant. Provide a detailed answer for the question that fully integrates all the obligations and best practices from the given passages. Ensure your response is cohesive and directly addresses the question. Synthesize the information from all passages into a single, unified answer.}

\subsection*{Prompt for obligations in the context (VRR)}
\label{app:obligations_context_prompt}
\textit{You are a regulatory compliance assistant. Your task is to provide a brief but concise and detailed answer to the Question, ensuring that all Obligations are fully addressed. Directly integrate each obligation into the response, ensuring no obligation is missed or implied. Avoid adding information beyond what is explicitly stated in the Obligations, and cite specific rules when necessary. Use the exact terminology and structure from the obligations where applicable, to ensure high alignment and logical consistency. Focus solely on the provided obligations to craft a response that is well-structured, concise, and free of contradictions.}

\subsection*{Prompt for inserting obligations (VRR)}
\label{app:obligations_insertion_prompt}
\textit{You are a regulatory compliance assistant. Your task is to integrate the following Obligations that are missing from the Answer. You may change sentences or add new ones to cover all Obligations. Avoid adding changes or sentences that contradict the Answer and/or the Obligations.}

\subsection*{Prompt that rewrites an obligation (LOC)}
\label{app:loc_prompt}
\textit{You are a regulatory compliance assistant. Your task is to construct a brief but concise response that addresses the Question by focusing exclusively on the specified Obligation. Ensure your response clearly identifies and explains the obligation, including any relevant conditions or restrictions. Avoid addressing unrelated aspects of the Question, and limit your response strictly to what is explicitly stated in the provided passage.}

\section{Detailed experiments for VRR}
Table~\ref{tab:vrr_results} shows the progression of RePASs throughout the execution of the VRR algorithm. The Verification step leads to an increase in all metrics. Obligation Refinement  (`Ref.\ Obl.') alone does not lead to an increased score, Contradiction Refinement (`Ref.\ Contr.') is necessary. Even though Obligation Coverage ('Obl.') increases at the expense of the Entailment ('Ent.') score, RePASs improves overall.

\begin{table}[h]
\setlength{\tabcolsep}{4pt}
\centering
\begin{tabular}{|c|c|c|c|c|}
\hline
\textbf{Step}       & \textbf{RePASs} & \textbf{Obl.} & \textbf{Ent.} & \textbf{Con.} \\ \hline
Preprocessing      & 0.506          & 0.246                      & 0.408              & 0.136                 \\ \hline
Verify              & 0.611          & 0.389                      & 0.527              & 0.083                 \\ \hline
Ref. Contr. 1             & 0.638          & 0.389                      & \textbf{0.554}              & 0.030                 \\ \hline
Ref. Obl. 1            & 0.634          & 0.465                      & 0.490              & 0.053                 \\ \hline
Ref. Contr. 2            & 0.643          & 0.464                      & 0.497              & 0.032                 \\ \hline
Ref. Obl. 2            & 0.637          & 0.496                      & 0.464              & 0.049                 \\ \hline
Ref. Contr. 3           & 0.643          & 0.494                      & 0.467              & \textbf{0.030}                 \\ \hline
Ref. Obl. 3           & 0.642          & 0.527                      & 0.446              & 0.046                 \\ \hline
Ref. Contr. 4       & \textbf{0.647}          & 0.525                      & 0.446              & 0.031                 \\ \hline
Ref. Obl. 4           & 0.641          & 0.538                      & 0.430              & 0.045                 \\ \hline
\end{tabular}
\caption{RePASs progress during VRR execution.}
\label{tab:vrr_results}
\end{table}

\end{document}